\documentclass[11pt,a4paper]{article}
\usepackage[hyperref]{rocling2026}
\usepackage{times}
\usepackage{latexsym}

\usepackage{microtype}
\usepackage{graphicx}
\usepackage{siunitx}
\usepackage{placeins}
\usepackage{fontawesome6}
\roclingfinalcopy 

\title{\textsc{SinBrief}: A Hybrid Framework for Abstractive Text Summarisation of Sinhala Legal Documents}

\author{Minduli Lasandi$^\spadesuit$ \and Nevidu Jayatilleke$^\clubsuit$ \\
  $^\spadesuit$School of Computing, 
  Informatics Institute of Technology, 
  Sri Lanka \\
  $^\clubsuit$Department of Computer Science \& Engineering, University of Moratuwa, Sri Lanka \\
  \texttt{minduli.20220374@iit.ac.lk, nevidu.25@cse.mrt.ac.lk}
  \small{
  }}

\date{}

\begin{document}
\maketitle


\begin{abstract}
Legal document summarisation in low-resource languages presents significant challenges due to the scarcity of annotated data and the complexity of domain-specific terminology. This paper presents \textsc{SinBrief}, a hybrid abstractive summarisation framework for Sinhala legal documents that does not require human-annotated training data. The proposed framework combines domain-aware word graph construction with neural sentence scoring to generate abstractive summaries from Sinhala legal text. Five sentence scoring models are evaluated within the framework: \texttt{mBert}, \texttt{Llama 3.1}, \texttt{Falcon 7B}, \texttt{Laser}, and a continually pre-trained \texttt{Llama} model domain-adapted to Sinhala legal text. The framework is evaluated on a Sinhala legal corpus using reference-free metrics, including Coverage, Density, Compression Ratio, SummaC, and Self-BertScore. Experimental results demonstrate that \textsc{SinBrief} produces summaries with lower lexical overlap than extractive baselines while maintaining factual consistency, demonstrating the viability of hybrid, largely annotation-free abstractive summarisation for low-resource legal NLP tasks.
\end{abstract}



\section{Introduction}
Legal documents are often lengthy, complex, and linguistically dense, making them difficult for the general public to interpret efficiently \cite{jayatilleke2025hybrid}. In Sri Lanka, where legal records are predominantly written in Sinhala, the absence of automated summarisation tools further limits accessibility and information retrieval \cite{jayawardane2022automatic}. As a result, legal practitioners, researchers, and citizens must manually navigate extensive textual content to extract relevant information, which can be time-consuming and cognitively demanding.

Automatic abstractive text summarisation offers a promising solution by generating concise, coherent summaries that capture the semantic essence by producing novel sentences that capture the semantic meaning of a source document \cite{10795848}. Unlike extractive summarisation methods, which select sentences directly from the original text \cite{azam2025current}, abstractive approaches enable deeper semantic compression and improved readability. These properties are particularly beneficial in legal contexts, where extracting sentences may preserve redundancy and structural complexity rather than improving clarity.

While significant progress has been achieved in abstractive summarisation for high-resource languages such as English, limited attention has been given to low-resource languages due to the lack of resources \cite{munaf2024low}. The scarcity of large-scale annotated datasets and domain-specific corpora poses substantial challenges for supervised learning approaches, particularly in specialised domains such as legal documentation \cite{ariai2025natural}. In low-resource settings, methods that minimise reliance on human annotation offer a promising alternative, as they eliminate the dependency on parallel document–summary datasets while still enabling semantic content compression \cite{nikolov2020abstractive}. However, research on abstractive summarisation for Sinhala legal texts remains largely unexplored.

To address this gap, this research proposes \textsc{SinBrief}, a hybrid abstractive text summarisation framework\footnote{\url{https://github.com/Minduli-Lasandi/SinBrief}} for Sinhala legal documents. The framework is largely annotation-free, with the core summarisation pipeline operating without supervision and the scoring component employing weak supervision through constraint-guided label construction, requiring no human annotation. The proposed approach incorporates domain-aware word graph construction and sentence scoring to generate coherent and informative summaries.

\section{Existing Work}

Text summarisation has been widely studied in Natural Language Processing, with research spanning extractive, abstractive, supervised, and unsupervised methodologies. These efforts have driven significant methodological advancements across multiple languages and domains.

\subsection{BenSumm}

Bensumm is described as the first unsupervised abstractive summarisation framework for Bengali documents \cite{chowdhury-etal-2021-unsupervised}. This model adopts a graph-based, single-document approach that relies on POS tagging and a pre-trained language model named \texttt{ULMFiT} \cite{howard-ruder-2018-universal} embeddings to perform sentence clustering, word-graph construction and sentence fusion. By generating and ranking candidate paths using a ranking strategy based on the work of \citet{boudin-morin-2013-keyphrase} within cluster-specific word graphs, the system produces abstractive summaries without requiring parallel training data.

In addition to the model, the authors released a human-annotated benchmark dataset consisting of 139 document–summary pairs, characterised by a high degree of abstractiveness. Experimental results demonstrated that BenSumm outperformed established unsupervised extractive baselines such as LexRank \cite{erkan2004lexrank} and TextRank \cite{mihalcea-tarau-2004-textrank} in terms of ROUGE \cite{lin-2004-rouge} scores and human evaluation metrics. 

\textsc{SinBrief} adopts BenSumm's general pipeline of sentence clustering, word graph construction, and graph-based candidate generation, but introduces three key novelties: (i) domain-aware edge weighting that assigns higher weights to edges involving Sinhala legal keywords; (ii) a weakly supervised neural sentence scoring model that replaces BenSumm's fixed ranking strategy with a trained binary classifier for legal sentence quality; and (iii) continual pre-training of Llama 3.1 8B on the Sinhala legal corpus to produce a domain-adapted backbone.


\subsection{Summarisation of Legal Texts}
The summarisation of legal texts has become increasingly important due to their complexity and length, and researchers have developed a variety of extractive and abstractive methods leveraging advanced NLP techniques and large language models to address these challenges.

\citet{10677065} explored extractive and abstractive summarisation of legal documents using transformer models and large language models such as \texttt{GPT-4} \cite{achiam2023gpt} and \texttt{Llama-2} \cite{touvron2023llama}, supporting multilingual output and applications like chatbots. \citet{grover2003automatic} applied summarisation to legal judgments by classifying sentences according to their rhetorical roles, using NLP tools and linguistic features to guide summary extraction, while \citet{chheda-etal-2025-extract} also introduced a rhetorical role-based Extract-Explain-Abstract framework for generating abstractive summaries in low-resource legal settings. 

\citet{hachey2005automatic} explored machine learning approaches, including Naïve Bayes and maximum entropy, for extractive summarisation of legal texts, leveraging rhetorical status information to score sentences and structure summaries.

\citet{jayatilleke2025hybrid} proposed a hybrid approach for patent summarisation, combining LexRank for extractive sentence selection with a \texttt{Bart} model \cite{lewis-etal-2020-bart} fine-tuned via LoRA \cite{hu2022lora} for abstractive summary generation and domain generalisation across patent fields. Other research includes the \textit{PRODIGIT} system by \citet{pont2023legal}, which uses GPT-4 to generate issue-based abstractive summaries of Italian tax law rulings, the fuzzy logic-based headnote generation approach by \citet{megala2014feature}, the fine-tuning of Bart and Pegasus \cite{zhang2020pegasus} for legal summarisation by \citet{kasar2025enhancing} and the summarisation of legal judgments by \citet{grover-etal-2003-summarising} through classifying sentences based on their argumentative roles using linguistic features.

\subsection{Unsupervised Summarisation Approaches}

Unsupervised summarisation methods aim to automatically generate concise, coherent summaries without requiring parallel document-summary datasets. By leveraging techniques ranging from graph-based representations to neural architectures and hybrid frameworks, these approaches enable effective summarisation in domains where annotated data is rare.

\citet{nayeem-etal-2018-abstractive} proposed an unsupervised abstractive summarisation model based on paraphrastic sentence fusion. The approach combines word-graph generation with semantic ranking to produce more informative summaries. \citet{liu-etal-2015-toward} proposed an AMR-based abstractive summarisation framework that generates summaries by transforming semantic graphs, while \citet{dohare-etal-2018-unsupervised} introduced an unsupervised approach that improves document-level graph construction and subgraph extraction for summary generation. \citet{ramirez-orta-milios-2021-unsupervised} introduced an unsupervised extractive method that ranks sentences using graph centrality over pre-trained sentence embeddings to capture semantic similarity. These approaches demonstrate the progression from simple word-graph and AMR-based methods toward more sophisticated document-level graph construction and embedding-based ranking.

\citet{chu2019meansum} proposed MeanSum, an end-to-end neural model for unsupervised abstractive summarisation that generates summaries by decoding the mean representation of input documents. \citet{zhou-rush-2019-simple} proposed an unsupervised summarisation approach based on contextual matching using pre-trained language models, while \citet{coavoux-etal-2019-unsupervised} introduced an aspect-aware neural model that generates summaries by clustering document representations into semantic facets.

Other research includes \citet{wu-etal-2022-unsupervised}, which proposed an unsupervised abstractive summarisation framework to identify salient content using frequency-based semantic units, \citet{tang2023topiccat}, which applies a topic-guided co-attention transformer for extreme multimodal summarisation, \citet{song-etal-2022-unsupervised}, an unsupervised abstractive model leveraging Wasserstein distance for opinion summarisation, and \citet{zhao2020summpip}, a graph-based method that combines sentence representations and spectral clustering for multi-document summarisation. These methods show that effective abstractive summarisation can be achieved without massive pre-trained models, making them suitable for low-resource domains.



\section{Methodology}

The proposed framework follows an hybrid abstractive summarisation approach designed for Sinhala legal texts. The method begins with sentence extraction and clustering, followed by graph-based representation to capture structural and semantic relationships within the document. Abstractive candidate sentences are then generated using a word graph-based fusion mechanism. To ensure the quality of the generated summaries, a scoring component is employed to rank candidate sentences. The top-ranked candidates are finally selected and combined to produce the final summary.


\subsection{Data}
This study utilises the \texttt{SinhaLegal} dataset, comprising 1206 legal documents of Acts and Bills \cite{lasandi-jayatilleke-2026-sinhalegal}. The dataset was partitioned into 80\% training and 20\% testing using a fixed random seed (42), resulting in 964 training documents and 242 testing documents.

\subsection{Sentence Extraction and Clustering}

The input document is segmented into sentences by splitting on punctuation marks, specifically periods, exclamation marks, question marks, and semicolons. Sentences shorter than 20 characters are discarded to eliminate incomplete fragments.

The retained sentences are then encoded into dense vector representations for semantic clustering. The choice of encoder varies depending on the scoring model used in the pipeline. For mBert-based and the continually pre-trained Llama variants, sentence representations are obtained from transformer models, while \texttt{Laser} employs its dedicated multilingual \textit{encode\_sentences()} API directly. For \texttt{mBert} \cite{devlin-etal-2019-bert}, the [CLS] token embedding from the final hidden layer serves as the sentence representation. 

For \texttt{Llama 3.1 8B} \cite{grattafiori2024llama} and \texttt{Falcon 7B} \cite{almazrouei2023falcon}, which are causal decoder-only models with no [CLS] token, the hidden state of the last non-padding token from the final layer is used instead. Both large models are quantised to 4-bit NF4 precision via BitsAndBytes to enable inference within GPU memory constraints. For \texttt{Laser} \cite{artetxe2019massively}, fixed-size 1024-dimensional embeddings are obtained directly from the frozen multilingual encoder without any token-level extraction. 

Regardless of the encoder used, sentence representations are grouped into semantically coherent clusters using agglomerative clustering with average linkage and cosine similarity as the distance metric. The number of clusters is set to $k = \min(3, \lfloor N/5 \rfloor)$, where $N$ is the total number of extracted sentences, capping the cluster count at 3, while reducing it for shorter documents with fewer than 15 sentences.

\subsection{Word Graph Construction}

For each sentence cluster, a directed word graph is constructed to capture co-occurrence relationships between words. Sentences within a cluster are first tokenised and POS-tagged using the \textit{Sinling} library, which provides Sinhala-specific tokenisation and morphological analysis.  Only tokens belonging to content-bearing POS categories (nouns, verbs, auxiliary verbs, proper nouns, and adjectives) are retained for graph construction. Each node in the graph represents a (word, POS tag) tuple, with directed edges connecting consecutive word pairs within a sentence. A START and END node are included to delimit valid sentence boundaries during candidate generation.

Edge weights are assigned using a domain-aware weighting scheme tailored to the legal domain. An edge receives a weight of 2.0 if either of its constituent nodes contains a term from a predefined vocabulary of Sinhala legal keywords. All remaining edges receive a default weight of 1.0. 
These terms were taken from a publicly available website\footnote{\url{https://talkpal.ai/culture/what-are-the-legal-terms-in-sinhala/}} and manually modified to include additional terms in the dataset. These words were validated and discussed in Appendix \ref{sec:appendix_keyword_validation}.

\begin{figure}[h]
  \centering
  \includegraphics[width=0.95\columnwidth,]{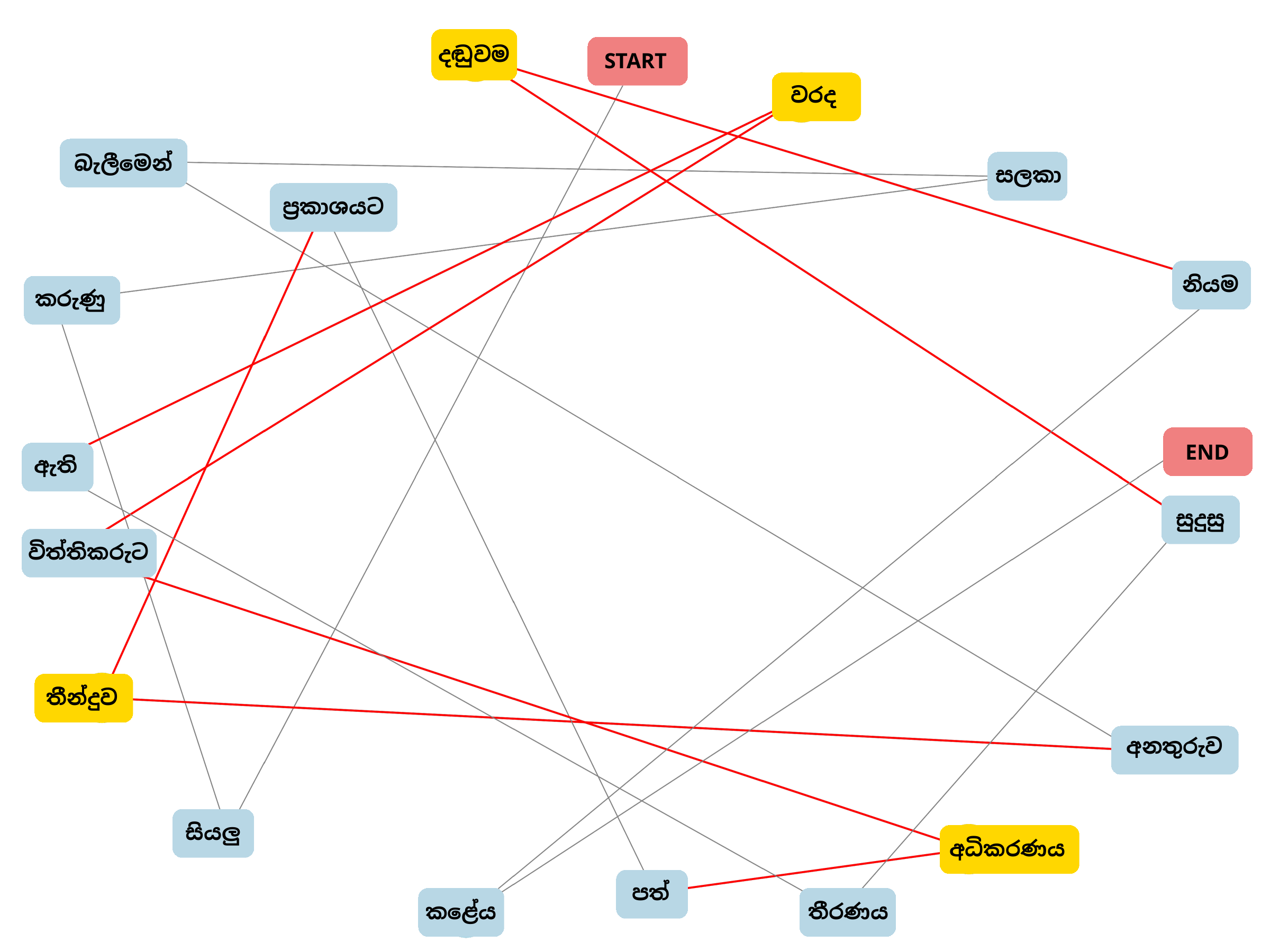}
    \caption{Example of a word graph. START and END nodes are depicted in red. Yellow nodes represent legal keywords, with red edges indicating their connections. Blue nodes represent regular words with grey edges indicating their connections.}
  \label{fig:word_graph_example}
\end{figure}

Given the morphological complexity of Sinhala, POS tagging does not always yield sufficient tagged tokens. When fewer than three content-bearing tokens are produced for a sentence, a fallback mechanism retains all tokens longer than two characters and assigns them a default NOUN tag, preserving graph connectivity and ensuring no sentence is entirely excluded from the graph structure. Figure \ref{fig:word_graph_example} depicts the word graph generated for a short document.

\subsection{Abstractive Candidate Generation}

For each cluster, abstractive sentence candidates are generated by performing weighted random walks over the word graph. Starting from the START node, each walk probabilistically selects the next node proportional to edge weights, biasing traversal toward paths containing legal keywords. Each walk produces a candidate of between 8 and 25 words, with up to 8 candidates generated per cluster.  

If the random walk procedure yields insufficient candidates due to sparse graphs, a fallback mechanism constructs candidates by ranking nodes by cumulative edge weight and sampling from the top-weighted words. All candidates are then subjected to quality filtering, where sequences shorter than 20 characters, those with less than 60\% unique words, or those containing any single word repeated three or more times are discarded.

\subsection{Candidate Scoring}

Each abstractive candidate generated from the word graph is evaluated by a weakly supervised sentence scoring model, which assigns a quality score in the range [0,1] reflecting how well the candidate captures meaningful legal content.Candidates are ranked per cluster, and the top two scoring candidates from each cluster are selected for inclusion in the final summary.

\subsubsection{Model Selection}
Four models were initially evaluated as candidate sentence scorers within the proposed framework: \texttt{mBert, Llama 3.1 8B, Falcon 7B}, and \texttt{Laser}. \texttt{mBert} was selected as a natural multilingual baseline given its established performance on low-resource languages. \texttt{Laser} was included as a lightweight multilingual alternative specifically designed for cross-lingual sentence representation. 

\texttt{Llama 3.1} 8B and \texttt{Falcon 7B} were selected based on perplexity evaluation on the Sinhala legal corpus, where they achieved the lowest perplexity scores of 2.55 and 2.61, respectively, indicating stronger language modelling capability on the target domain compared to other large language models considered \cite{lasandi-jayatilleke-2026-sinhalegal}.

Based on the summary quality evaluation results discussed in Section~\ref{sec:evaluation}, \texttt{Llama 3.1 8B} produced the highest quality summaries among the four models, and therefore, it was selected for continual pre-training (CPT) on the Sinhala legal domain to improve its capacity further. This adds a 5th sentence scorer, the continually pre-trained \texttt{Llama 3.1}, which was then evaluated alongside the original 4 models.

\subsubsection{Continual Pre-Training of Llama 3.1}

To adapt \texttt{Llama 3.1 8B} to the Sinhala legal domain, continual pre-training was performed on the training section of the \texttt{SinhaLegal} dataset. The training set consisted of 964 legal documents.  The resulting domain-adapted model, \texttt{SinBrief-Legal-Llama 3.1}\footnote{\url{https://huggingface.co/Minduli-Lasandi/SinBrief-Legal-Llama3.1}}, did not surpass the base \texttt{Llama 3.1} in scoring performance, a finding reported as a negative result. Full implementation details are provided in Appendix~\ref{Sec: appendix_CPTLlama}.

\subsubsection{Training sentence scorers}

A sentence scoring model was trained for each of the five encoder backbones: \texttt{mBert, Llama 3.1 8B, Falcon 7B, Laser}, and \texttt{CPT Llama 3.1}. This was done using a weakly supervised binary classification objective. The full dataset of 1,206 Sinhala legal documents was first partitioned into 80\% training (964 documents) and 20\% test (242 documents) sets, with the test set held out entirely for final evaluation. The training data was then further split into 80\% for model training and 20\% for validation using stratified sampling to maintain class balance, yielding 38,920 labelled sentence samples in total for scorer training.

Positive samples (label = 1) were constructed using four strategies applied to the 964 training documents. First, the opening sentence of each document was extracted, as it typically introduces the legal matter at hand. Second, the closing sentence was extracted, as it generally contains the verdict or conclusion. Third, any sentence containing at least one of the 19 predefined Sinhala legal keywords was included. Fourth, sentences exceeding 50 characters were added up to a cap of 5,000 positive samples, as longer sentences tend to carry more substantive legal content. 

Negative samples (label = 0) were synthetically generated by applying one of four corruption strategies to the positive sentences: word shuffling, which destroys grammatical structure; random word deletion, which removes approximately 30\% of tokens; random word replacement, which substitutes tokens with randomly sampled vocabulary words; and sentence truncation, which retains only the first half of the token sequence. Each positive sentence was corrupted using one randomly selected method, producing an equal number of negative samples and resulting in a balanced dataset. The final training dataset consisted of 38,920 samples in total, which was split 50\% positive and 50\% negative.

\begin{figure}[h]
  \centering
  \includegraphics[width=\columnwidth,]{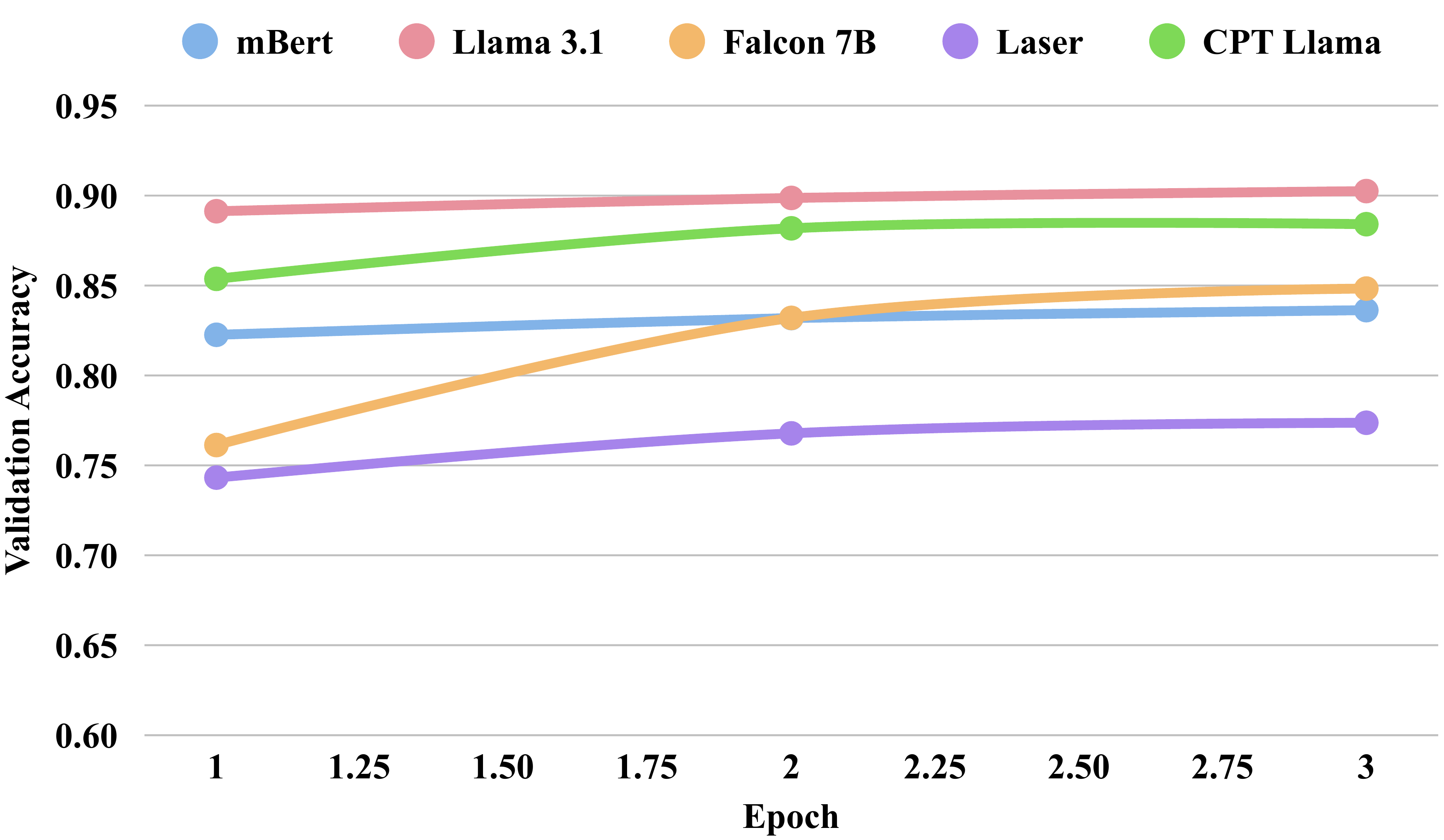}
  \caption{Validation accuracy of the sentence scorers}
  \label{fig:validation_accuracy}
\end{figure}

Each scorer consists of the respective encoder backbone followed by a shared three-layer classification head: a fully connected layer projecting from the encoder's hidden size to 256 dimensions, followed by a second layer reducing to 128 dimensions, and a final layer producing a scalar score in the range [0,1] through a sigmoid activation. ReLU activations and dropout (p=0.3) are applied after each intermediate layer. For \texttt{mBert}, the full model including the backbone was fine-tuned end-to-end, while for the large decoder-only models (\texttt{Llama 3.1, Falcon 7B}, and \texttt{CPT Llama}), only the classification head was trained with the quantised backbone kept frozen, as full fine-tuning was computationally infeasible. For \texttt{Laser}, embeddings were pre-computed using the frozen encoder, and only the classification head was trained, making it the most lightweight of the five configurations.

All models were trained for 3 epochs with a batch size of 16, using the AdamW optimiser with a learning rate of \num{2e-5}, weight decay of 0.01, and a linear warmup over 100 steps. Binary cross-entropy was used as the loss function, with gradient clipping at 1.0 to ensure training stability.

Figure~\ref{fig:validation_accuracy} shows the validation accuracy across epochs for all five scorers. \texttt{Llama 3.1} achieved the highest validation accuracy of 90.05\%, followed closely by \texttt{CPT Llama} at 88.37\%. \texttt{mBert} and \texttt{Falcon 7B} reached 83.83\% and 84.96\%, respectively, while \texttt{Laser} achieved 77.25\%. The consistently strong performance of \texttt{Llama 3.1} and \texttt{CPT Llama} suggests that larger decoder-only models may produce more discriminative sentence representations for the Sinhala legal domain. The Training curves of these models are discussed in Appendix \ref{Sec: appendix_Training curves}.

\subsection{Candidate Selection}

Following the scoring stage, all candidates generated for each cluster are ranked in descending order of their scorer output. The top two highest-scoring candidates per cluster are selected for inclusion in the final summary. With three clusters per document, this yields a maximum of six selected sentences. This per-cluster selection strategy ensures that the final summary draws content from semantically distinct regions of the document, preserving topical coverage rather than repeatedly selecting candidates from a single dominant theme.

Qualitative observation of the assigned scores further supports the quantitative findings. \texttt{Llama 3.1} and \texttt{CPT Llama} assigned scores of 0.9 and above to high-quality legal sentences, demonstrating strong discriminative ability. \texttt{Laser} produced scores of approximately 0.8, and  \texttt{Falcon 7B} produced moderate scores of approximately 0.6 and 0.7. \texttt{mBert} assigned the lowest scores overall, with most candidates receiving values near 0.5.

\begin{figure}[h]
  \centering
  \includegraphics[width=\columnwidth,]{Diagrams/Candidate_Scores.pdf}
  \caption{Scores given by the CPT Llama 3.1.}
  \label{fig:candidate_scores}
\end{figure}

Figure \ref{fig:candidate_scores} illustrates an example of candidate scoring by the \texttt{CPT Llama} model, where the two highest-scoring candidates (0.9080 and 0.8426) are correctly selected while lower-quality candidates receive substantially lower scores.

\subsection{Final Summary}

The top two highest-scoring candidates per cluster are combined to form the final summary. The combined text is then subject to a length control mechanism targeting a character range of 300 to 500 characters, which was determined to be appropriate for a concise legal summary. If it exceeds 500 characters, sentences are added incrementally until the target range is reached, with the final sentence truncated at the nearest word boundary if necessary to stay within the limit. Figure \ref{fig:sinhala_summary} shows an example of a generated summary. The summaries generated by all the 5 models are shown in Appendix \ref{Sec: appendix_final_summaries}.

\begin{figure}[h]
  \centering
  \includegraphics[width=\columnwidth,]{Diagrams/Final_Summary.pdf}
  \caption{Example of a generated summary}
  \label{fig:sinhala_summary}
\end{figure}

\section{Evaluation}
\label{sec:evaluation}

To evaluate the framework, experiments were conducted on the test set held out. This test set comprised 242 legal documents, representing 20\% of the full corpus. The framework was evaluated under five configurations corresponding to each of the five sentence scoring models and compared against five baseline summarisation systems.

\subsection{Selection of Evaluation Metrics}
Since no reference summaries existed for the \texttt{SinhaLegal} dataset, standard reference-based metrics such as ROUGE cannot be applied. Instead, a set of reference-free evaluation metrics was selected to assess the quality of the generated summaries from multiple perspectives.

Coverage and Density measure the degree to which the summary content is drawn from the source document \cite{grusky-etal-2018-newsroom}. Coverage is the proportion of summary tokens that appear in contiguous extractive fragments shared with the source. Density is the average squared length of extractive fragments, indicating the length and coherence of copied spans. An abstractive summary is expected to depict lower coverage and density values compared to an extractive summary.

Compression Ratio is defined as the ratio of the summary length in words to the source document length in words \cite{liu-etal-2022-reference}. It quantifies how concisely the framework summarises the original document. Self-BertScore measures the semantic similarity between the generated summary and the source document using contextual embeddings, without requiring a reference summary \cite{zhang2019bertscore}. Following the approach used for low-resource languages, BertScore is computed using \texttt{mBert} as the backbone model.

SummaC is a reference-free factual consistency metric that evaluates whether the content of the summary is entailed by the source document \cite{laban2022summac}. It uses a natural language inference model to score sentence-level consistency between the summary and the source, with higher scores indicating greater factual faithfulness. Specifically, SummaCConv internally employs the VitC (VitaminC) NLI model for sentence-level entailment scoring.

\subsection{Baseline Systems}

Five baseline summarisation systems were evaluated on the test set for comparison with the proposed framework. Three extractive baselines and two abstractive baselines were included. Lead-3 serves as the simplest extractive baseline, selecting the first three sentences of each document as the summary \cite{see-etal-2017-get}. 

TextRank \cite{mihalcea-tarau-2004-textrank} is a graph-based extractive method that ranks sentences using a PageRank algorithm applied to a sentence similarity graph. Sentence embeddings were computed using \textit{paraphrase-multilingual-MiniLM-L12-v2} to support Sinhala text. PacSum \cite{zheng-lapata-2019-sentence} extends the TextRank approach by introducing directional edge weights in the sentence graph, assigning greater importance to forward-looking sentences.

The fine-tuned \texttt{mT5} model\footnote{\url{https://huggingface.co/Navanjana/Sinhala-Sumarization}} is an abstractive baseline fine-tuned on a Sinhala dataset. BenSumm \cite{chowdhury-etal-2021-unsupervised} is an abstractive graph-based summarisation framework originally proposed for Bengali. Since the original model relies on a Bengali-specific ULMFiT backbone, which cannot be applied to Sinhala, BenSumm was adapted by replacing the backbone with \texttt{mBert}.

Prior to finalising the baseline selection, zero-shot inference was attempted using \texttt{mT5} \cite{xue-etal-2021-mt5}, and \texttt{mBart} \cite{liu2020multilingual} with no fine-tuning. As shown in Figure \ref{fig:mt5_zeroshot}, both models produced degraded outputs containing special tokens such as \textit{$<$extra\_id\_0$>$}, fragmented Sinhala text, and mixed-script characters, rendering the summaries unusable.

\begin{figure}[h]
  \centering
  \includegraphics[width=\columnwidth]{Diagrams/mt5_Zeroshot_Error.pdf}
  \caption{Output from the zero-shot mT5 method.\\
  \scriptsize *Examples of characters that are not in Sinhala are enclosed in red boxes, while repeated words are highlighted with green underlining.}
  \label{fig:mt5_zeroshot}
\end{figure}

\subsection{Results and Discussion}

\begin{table*}[!t]
\centering
\small
\renewcommand{\arraystretch}{1.2}
\begin{tabular}{p{3cm} c c c c c}
\hline

\textbf{Model} 
& \textbf{Coverage} 
& \textbf{Density} 
& \textbf{Comp. Ratio} 
& \textbf{SummaC} 
& \textbf{Self-BertScore} \\

\hline
 Lead-3 & 1.000 & 8.153 & 0.129 & 0.488 & 0.850 \\
 PacSum & 1.000 & 8.150 & 0.265 & 0.481 & \textbf{0.896} \\
 TextRank & 1.000 & 7.605 & 0.240 & 0.648 & \underline{0.859} \\

 Fine-tuned mT5 & 0.855 & 3.989 & 0.099 & 0.448 & 0.802 \\
 Bensumm & 0.913 & 4.205 & 0.047 & 0.452 & 0.793 \\

\hline
\multicolumn{6}{c}{\textbf{Sentence scorers - \textsc{SinBrief} }} \\
\hline

 mBert & 0.958 & \textbf{2.924} & 0.084 & 0.584 & 0.798 \\
 Llama 3.1 & \textbf{0.956} & 3.091 & 0.089 & \textbf{0.752} & 0.803 \\
 Falcon-7B & \underline{0.957} & \underline{2.970} & 0.088 & \underline{0.740} & 0.809 \\
 Laser & 0.959 & 3.125 & 0.084 & 0.471 & 0.809 \\
 CPT Llama & 0.959 & 3.125 & 0.090 & 0.591 & 0.804 \\

\hline
\end{tabular}
\caption{Evaluation of existing systems and \textsc{SinBrief} sentence scorers using the \textsc{SinhaLegal} corpus.\\ \scriptsize
\textbf{Bold:} indicates best performance and \underline{Underline:} indicates the second best. Values shown in the figure are rounded to three decimal places.}
\label{tab:results}
\end{table*}

The results in Table \ref{tab:results} demonstrate that \textsc{SinBrief} successfully generates abstractive summaries of Sinhala legal documents, with all five scoring configurations achieving substantially lower Coverage (0.957–0.959) and Density (2.924–3.125) scores than the extractive baselines (Coverage: 1.000, Density: 7.605–8.153).  This indicates that the word graph-based candidate generation produces sentences with substantially lower lexical overlap with the source compared to extractive baselines. The fine-tuned mT5 and adapted BenSumm also show reduced Coverage and Density scores (0.855–0.913, 3.989–4.205), though remaining higher than \textsc{SinBrief}, indicating that \textsc{SinBrief} generates more novel text than any of the baseline systems.

In terms of factual consistency, \texttt{Llama 3.1} achieves the highest SummaC score among all evaluated systems (0.752), outperforming not only the other \textsc{SinBrief} models but also all five baseline systems, including TextRank, which achieves the highest SummaC among baselines (0.648). \texttt{Falcon 7B} follows closely (0.740), suggesting that large decoder-only language models provide more effective scoring signals for selecting factually consistent candidates.


Two extractive baselines (Lead-3 and PacSum) achieved unexpectedly low SummaC scores despite directly extracting from the source. This is attributed to the granularity mismatch identified by \citet{laban2022summac}, where sentence-level NLI models struggle with long, complex sentences that differ significantly from the news corpora on which SummaC was benchmarked \cite{koreeda-manning-2021-contractnli-dataset}, with declining reliability further documented for long documents \cite{mujahid2026stress}. As Sinhala legal sentences are particularly long and morphologically complex, the shorter graph-generated candidates of \textsc{SinBrief} are more readily verified, and the reported SummaC scores are therefore interpreted as relative indicators.

Regarding the semantic similarity, all \textsc{SinBrief} models achieved competitive self-BertScore results (0.798–0.809) compared to the abstractive baselines fine-tuned mT5 (0.802) and BenSumm (0.793), and approached the scores of extractive baselines such as Lead-3 (0.850) and TextRank (0.859). This demonstrates that despite lower lexical overlap with the source, \textsc{SinBrief} preserves strong semantic faithfulness to the source document. Among \textsc{SinBrief} models, \texttt{Laser} and \texttt{Falcon 7B} achieve the highest Self-BertScore (0.809), while \texttt{Llama 3.1} emerges as the strongest by achieving the best factual consistency while maintaining competitive abstractiveness and semantic similarity, consistent with its highest validation accuracy of 90.05\% during scorer training.

\section{Conclusion}

This paper presented \textsc{SinBrief}, a hybrid abstractive summarisation framework for Sinhala legal documents. The framework combines domain-aware word graph construction with sentence scoring to generate abstractive summaries without requiring annotated training data, addressing a critical gap in NLP resources for Sinhala. Five sentence scoring models were evaluated within the framework, and the evaluation against both extractive and abstractive baselines demonstrated that \textsc{SinBrief} produces more abstractive summaries while maintaining semantic faithfulness compared to existing approaches. Furthermore, continual pre-training of Llama 3.1 on the Sinhala legal corpus produced a domain-adapted language model specifically tailored to Sinhala legal text. However, CPT Llama did not surpass the base Llama 3.1 in sentence scoring performance, a finding reported as a negative result.

Future work includes incorporating grammatical constraints into the generation process to improve fluency, conducting human evaluation with native Sinhala speakers, and extending the framework to other low-resource languages.

\section*{Limitations}
\label{sec:Limitations}

Several limitations of the proposed framework should be acknowledged. The authors, who are native Sinhala speakers, manually identified grammatical errors in the generated summaries, arising from the random walk mechanism which connects words based on co-occurrence weights without grammatical constraints. As a result, the evaluation relies solely on quantitative metrics, highlighting the need for human evaluation in future work. 

The stochastic nature of the random walk further means that different summaries may be produced for the same document across runs; due to computational constraints of running large models such as \texttt{Llama 3.1} and \texttt{Falcon 7B}, multi-seed evaluation was not conducted. The reported validation accuracy of the sentence scorers reflects classification performance rather than ranking quality over graph-generated candidates, which is difficult to evaluate directly in this setting. 

Furthermore, the framework relies on Sinling for tokenisation and POS tagging, where frequent errors reduce graph construction quality. The Compression Ratio reflects the fixed length constraint rather than the model's inherent compression ability, and the individual contributions of domain-aware edge weighting and the legal keyword list remain unquantified due to the absence of ablation studies. Finally, the framework was evaluated on 1,206 Sinhala legal documents, which may not capture the full diversity of legal writing styles.

\section*{Acknowledgments}

We thank Ms. Sadini Jaburagoda, Attorney-at-Law, for reviewing the Sinhala legal keyword set and providing expert feedback on its relevance to Sri Lankan legal practice. Her evaluation helped assess the legal validity and domain coverage of the selected keywords and provided valuable insights into their applicability across different areas of legal discourse.


\bibliography{rocling2026}
\bibliographystyle{acl_natbib}

\appendix

\section{Continual Pre-Training of Llama 3.1}
\label{Sec: appendix_CPTLlama}

The documents were tokenised and segmented into fixed-length chunks of 512 tokens using a causal language modelling objective, where the model is trained to predict the next token given the preceding context.
 
Given the scale of \texttt{Llama 3.1 8B}, full fine-tuning was computationally infeasible. The model was therefore loaded in 4-bit NF4 quantisation via BitsAndBytes, and Parameter-Efficient Fine-Tuning was applied using LoRA. LoRA adapters were attached to all major projection layers, query, key, value, output, gate, up, and down projections, with a rank of
r=16 and scaling factor $\alpha$=32, resulting in a small fraction of trainable parameters relative to the full model size.

Training was conducted for 3 epochs with a batch size of 4, using the AdamW optimiser \cite{loshchilov2017decoupled} with a learning rate of \num{2e-5} and a linear warmup over 100 steps. After the completion of the CPT, the LoRA adapter weights were merged back into the base model and saved in bfloat16 precision.

\section{Training curves}
\label{Sec: appendix_Training curves}

\begin{figure}[h]
  \centering
  \includegraphics[width=\columnwidth,]{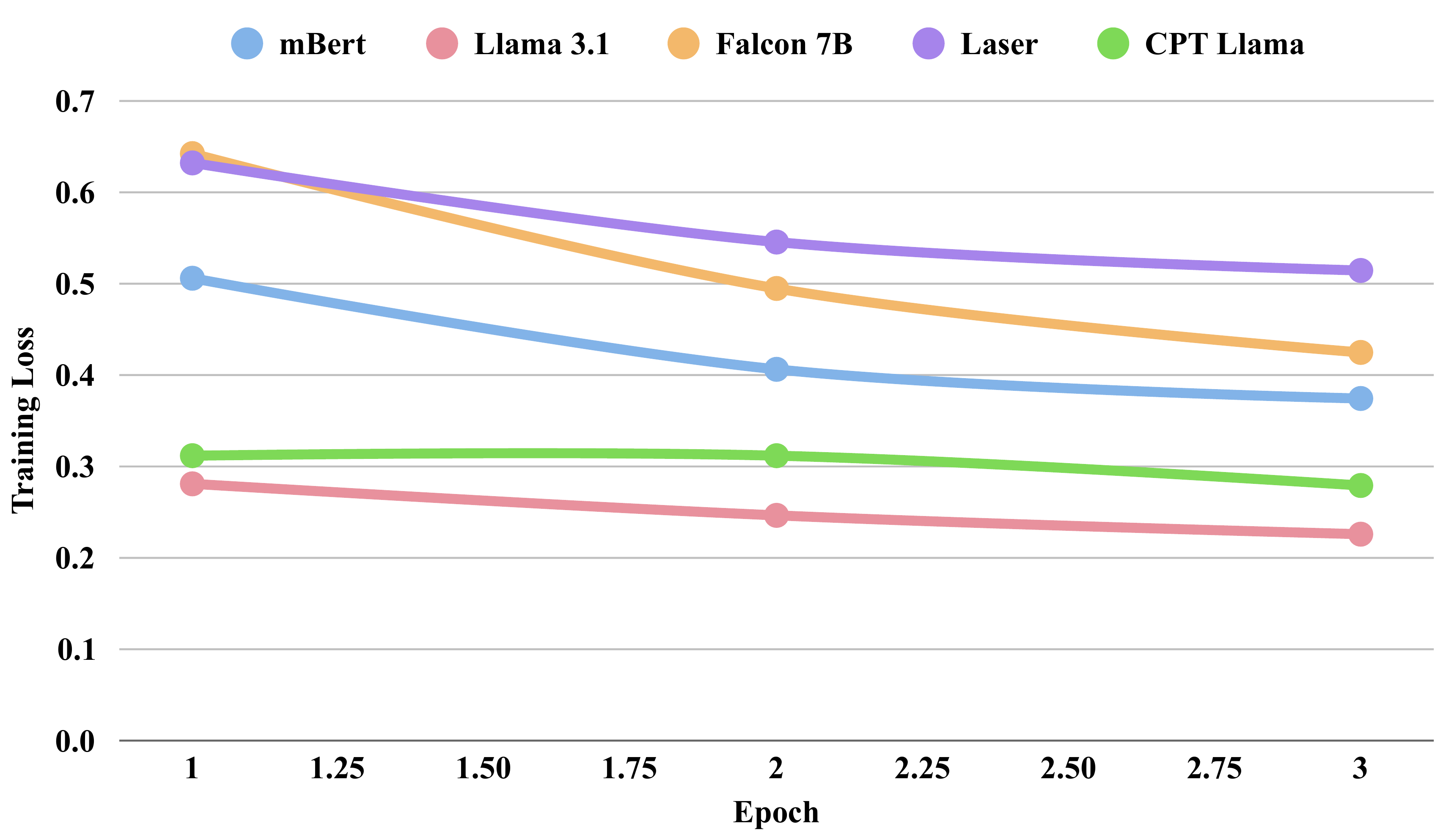}
  \caption{Training loss of the sentence scorers}
  \label{fig:training_loss}
\end{figure}

Figure~\ref{fig:training_loss} presents the training loss curves for all five sentence scoring models across three epochs. All five models demonstrate a consistent decrease in training loss across epochs, confirming stable convergence without signs of divergence or instability.

\texttt{Llama 3.1} achieves the lowest training loss across all epochs, starting at 0.281 in epoch 1 and decreasing to 0.2259 by epoch 3, consistent with its strongest validation accuracy of 90.05\%. \texttt{CPT Llama} follows closely, beginning at 0.3118 and converging to 0.2792, reflecting the benefit of domain adaptation through continual pre-training on Sinhala legal text. \texttt{mBert} shows steady convergence from 0.506 to 0.3744, while \texttt{Falcon 7B} decreases from 0.6424 to 0.4248. \texttt{Laser} records the highest training loss across all epochs, starting at 0.6322 and converging to 0.5145, consistent with its lower validation accuracy of 77.25\%.

\section{Final Summaries}
\label{Sec: appendix_final_summaries}

The Table \ref{tab:example_summaries}  presents the summaries generated by all five \textsc{SinBrief} scoring configurations for the same Sinhala legal document. These examples provide a qualitative illustration of how the choice of sentence scoring model influences candidate selection within the same underlying word graph-based generation pipeline. Since all five configurations share the same sentence extraction, clustering, word graph construction, and candidate generation steps, the differences observed across the summaries are attributable to the word graph walks and scoring models' ability to discriminate between high and low-quality candidates.

Despite sharing the same candidate pool, the selected sentences vary across configurations, reflecting the differences in scoring behaviour quantified in Table \ref{tab:example_summaries}. Models with higher SummaC scores, namely \texttt{Llama 3.1} and \texttt{Falcon 7B}, are expected to select candidates with stronger factual alignment to the source document, while models with lower discriminative ability, such as \texttt{Laser}, may select candidates that are semantically related but less factually grounded. Across all five summaries, grammatical errors, limited fluency, and reduced naturalness of Sinhala text are consistently observable, which is an inherent limitation of the graph-based random walk generation mechanism as discussed in Section \ref{sec:Limitations}. 

\begin{table}[!t]
\small
\renewcommand{\arraystretch}{1.4}
\begin{tabular}{p{1.5cm} l}
\hline
\textbf{Model} & \textbf{Generated Summary} \\
\hline
mBert & \includegraphics[width=5cm]{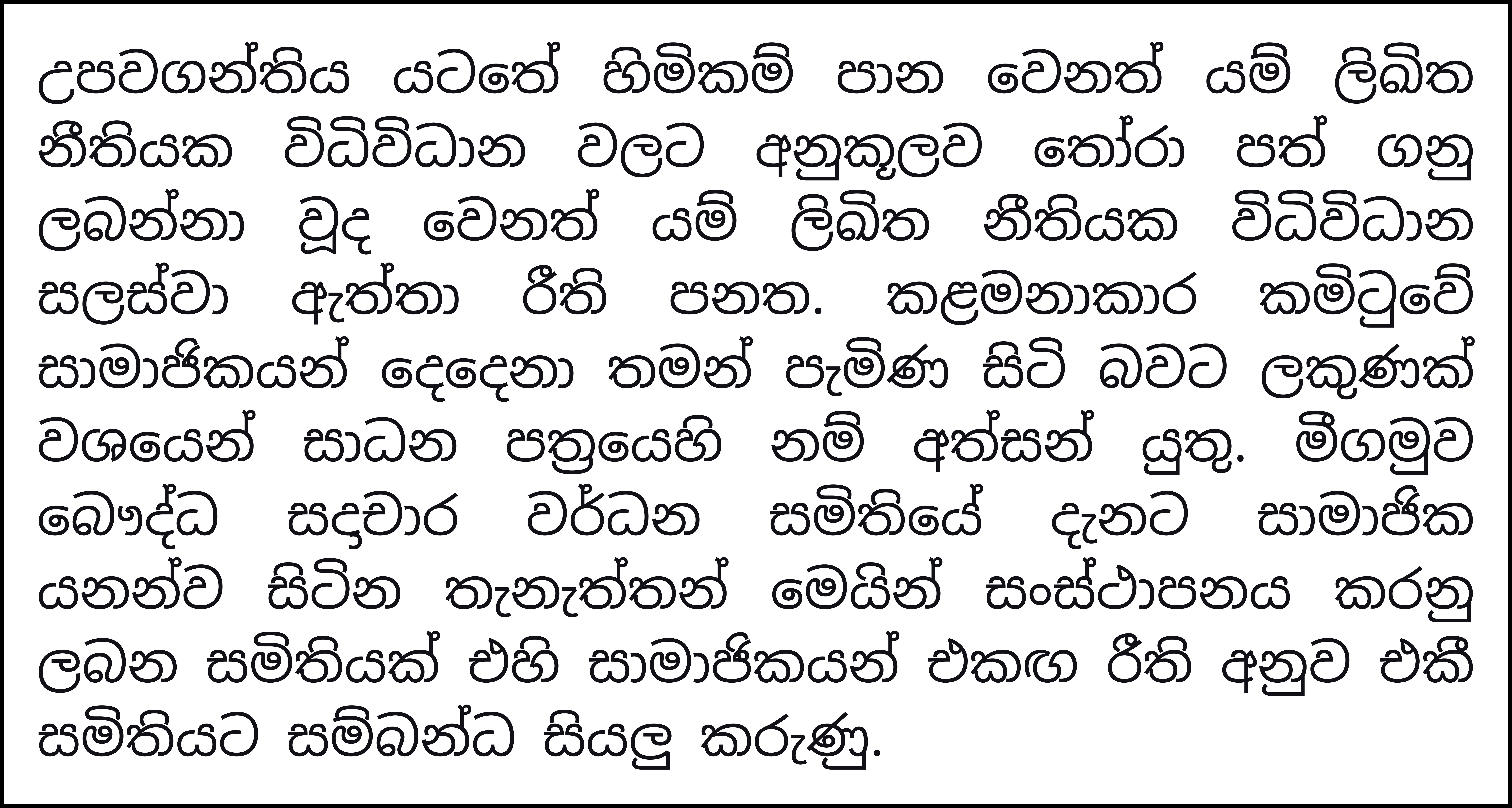} \\
Llama 3.1 & \includegraphics[width=5cm]{Diagrams/Final_Summary_Llama3.1.pdf} \\
Falcon 7B & \includegraphics[width=5cm]{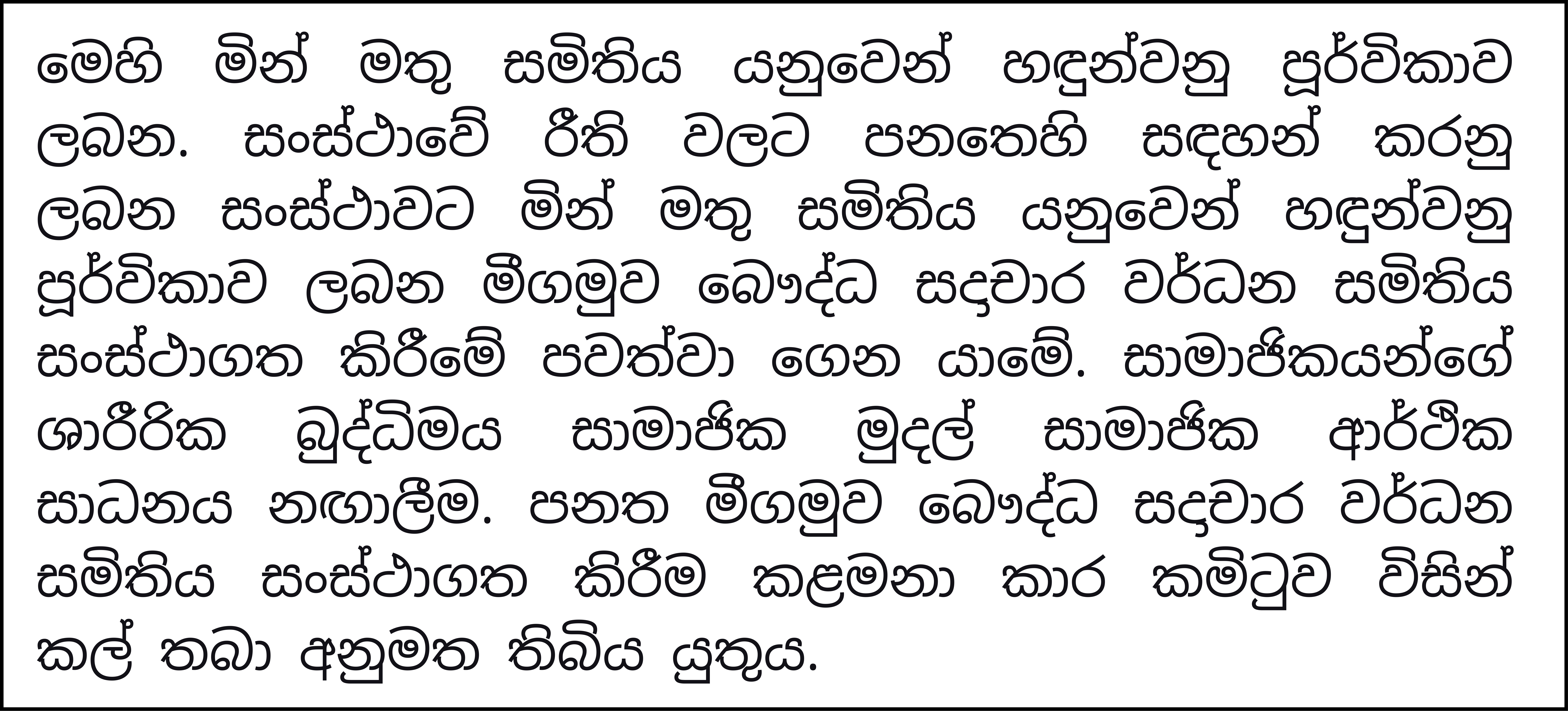}\\
Laser & \includegraphics[width=5cm]{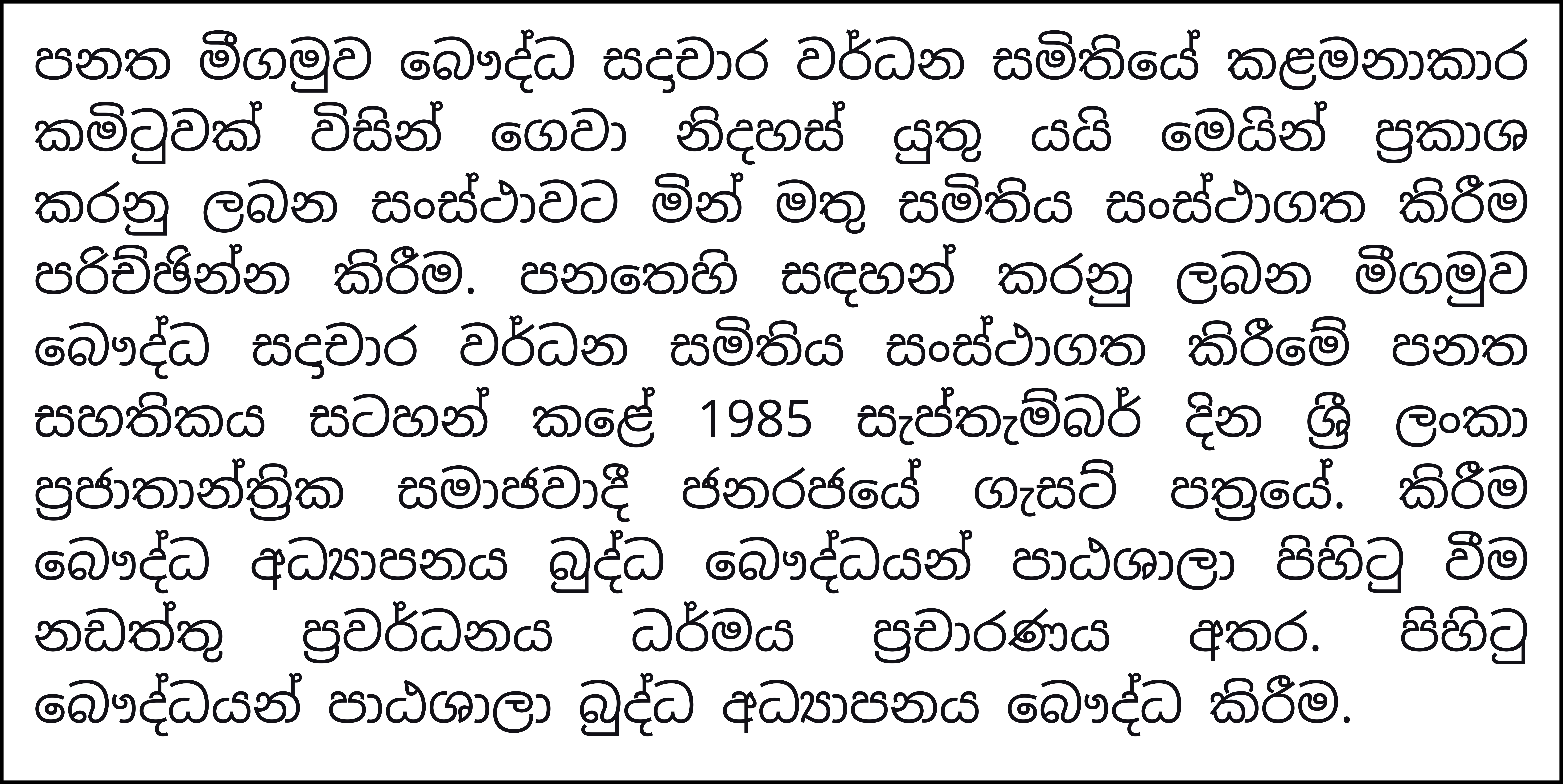} \\
CPT Llama & \includegraphics[width=5cm]{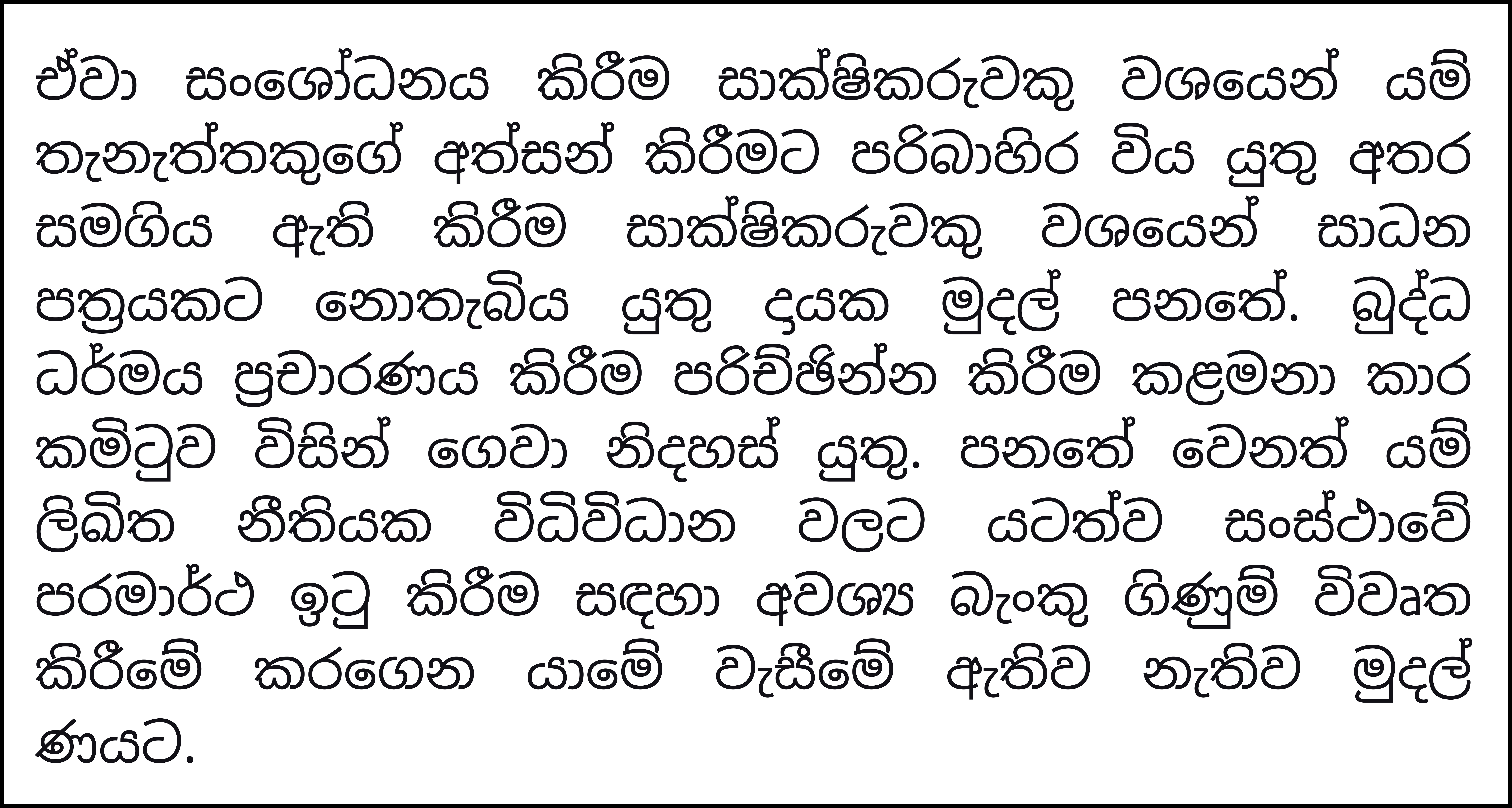} \\
\hline
\end{tabular}
\caption{Summaries generated by all five \textsc{SinBrief} models for the same source document. Document used was \textit{acts\_1985-09-17\_meegamuwa\_bauddha\_sadachara\_204} taken from the \texttt{SinhaLegal} Dataset.}

\label{tab:example_summaries}
\end{table}

\section{Legal Keyword Validation}
\label{sec:appendix_keyword_validation}

\subsection{Word Frequency Analysis}

To validate the domain relevance of the selected legal keywords, a corpus-wide frequency analysis was conducted across all 1,206 Sinhala legal documents. For each keyword, the document frequency, the number of documents in which the keyword appears at least once, and the total occurrence count across the entire corpus were recorded.  The results, presented in Table~\ref{tab:keyword_validation}, confirm that all 19 keywords appear across a substantial portion of the corpus, with high-frequency terms such as \hspace*{-3pt}\raisebox{-0.5ex}{ \includegraphics[height=1.3\fontcharht\font`\A]{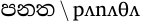}},  \hspace*{-3pt}\raisebox{-0.5ex}{ \includegraphics[height=1.3\fontcharht\font`\A]{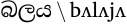}} and  \hspace*{-3pt}\raisebox{-0.5ex}{ \includegraphics[height=1.3\fontcharht\font`\A]{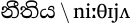}} appearing in over half of all documents. Lower-frequency terms such as \hspace*{-3pt}\raisebox{-0.5ex}{ \includegraphics[height=1.3\fontcharht\font`\A]{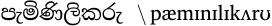}} and \hspace*{-3pt}\raisebox{-0.5ex}{ \includegraphics[height=1.3\fontcharht\font`\A]{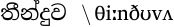}} appear less frequently but remain relevant to specific legal contexts such as litigation and sentencing. 

As observed in Table~\ref{tab:keyword_validation}, all 19 keywords record a total occurrence count exceeding 50 across the corpus, with the exception of \hspace*{-3pt}\raisebox{-0.5ex}{ \includegraphics[height=1.3\fontcharht\font`\A]{Sinhala_Words/si_word_pamilinikaru.pdf}} which records 39 total occurrences. Despite its relatively lower frequency, this term remains a legally significant keyword specific to litigation contexts. The remaining 18 keywords collectively account for thousands of occurrences with \hspace*{-3pt}\raisebox{-0.5ex}{ \includegraphics[height=1.3\fontcharht\font`\A]{Sinhala_Words/si_word_panatha.pdf}}  alone recording 24,293 occurrences across all 1,206 documents.

The keyword list was initially sourced from a publicly available Sinhala legal terminology reference\footnote{\url{https://talkpal.ai/culture/what-are-the-legal-terms-in-sinhala/}} and subsequently extended through manual inspection of the corpus to include additional domain-specific terms. The corpus frequency analysis presented in Table~\ref{tab:keyword_validation} further confirms the domain relevance of the selected keywords, demonstrating that all 19 terms are genuinely embedded in Sinhala legal writing across the corpus.

\begin{table}[!t]
\centering
\small
\renewcommand{\arraystretch}{1.2}
\begin{tabular}{l c c}
\hline
\textbf{Keyword} & \textbf{Doc. Frequency} & \textbf{Occurrences} \\
\hline
\hspace*{-3pt}\raisebox{-0.5ex}{ \includegraphics[height=1.3\fontcharht\font`\A]{Sinhala_Words/si_word_panatha.pdf}} & 1206 & 24293 \\
\hspace*{-3pt}\raisebox{-0.5ex}{ \includegraphics[height=1.3\fontcharht\font`\A]{Sinhala_Words/si_word_balaya.pdf}} & 773 & 3982 \\
\hspace*{-3pt}\raisebox{-0.5ex}{ \includegraphics[height=1.3\fontcharht\font`\A]{Sinhala_Words/si_word_niithiya.pdf}} & 715 & 2733 \\
\hspace*{-3pt}\raisebox{-0.5ex}{ \includegraphics[height=1.3\fontcharht\font`\A]{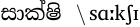}} & 487 & 1409 \\
\hspace*{-3pt}\raisebox{-0.5ex}{ \includegraphics[height=1.3\fontcharht\font`\A]{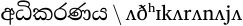}} & 354 & 2138 \\
\hspace*{-3pt}\raisebox{-0.5ex}{ \includegraphics[height=1.3\fontcharht\font`\A]{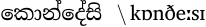}} & 321 & 912 \\
\hspace*{-3pt}\raisebox{-0.5ex}{ \includegraphics[height=1.3\fontcharht\font`\A]{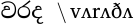}} & 299 & 4167 \\
\hspace*{-3pt}\raisebox{-0.5ex}{ \includegraphics[height=1.3\fontcharht\font`\A]{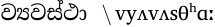}} & 257 & 1304 \\
\hspace*{-3pt}\raisebox{-0.5ex}{ \includegraphics[height=1.3\fontcharht\font`\A]{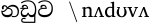}} & 214 & 830 \\
\hspace*{-3pt}\raisebox{-0.5ex}{ \includegraphics[height=1.3\fontcharht\font`\A]{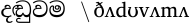}} & 183 & 455 \\
\hspace*{-3pt}\raisebox{-0.5ex}{ \includegraphics[height=1.3\fontcharht\font`\A]{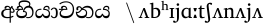}} & 130 & 698 \\
\hspace*{-3pt}\raisebox{-0.5ex}{ \includegraphics[height=1.3\fontcharht\font`\A]{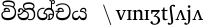}} & 119 & 715 \\
\hspace*{-3pt}\raisebox{-0.5ex}{ \includegraphics[height=1.3\fontcharht\font`\A]{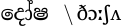}} & 99 & 121 \\
\hspace*{-3pt}\raisebox{-0.5ex}{ \includegraphics[height=1.3\fontcharht\font`\A]{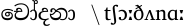}} & 72 & 225 \\
\hspace*{-3pt}\raisebox{-0.5ex}{ \includegraphics[height=1.3\fontcharht\font`\A]{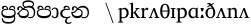}} & 52 & 80 \\
\hspace*{-3pt}\raisebox{-0.5ex}{ \includegraphics[height=1.3\fontcharht\font`\A]{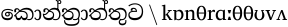}} & 46 & 51 \\
\hspace*{-3pt}\raisebox{-0.5ex}{ \includegraphics[height=1.3\fontcharht\font`\A]{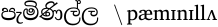}} & 37 & 83 \\
\hspace*{-3pt}\raisebox{-0.5ex}{ \includegraphics[height=1.3\fontcharht\font`\A]{Sinhala_Words/si_word_thiinduwa.pdf}} & 33 & 79 \\
\hspace*{-3pt}\raisebox{-0.5ex}{ \includegraphics[height=1.3\fontcharht\font`\A]{Sinhala_Words/si_word_pamilinikaru.pdf}} & 10 & 39 \\
\hline
\end{tabular}
\caption{Corpus frequency analysis of Sinhala legal keywords across 1,206 documents.}
\label{tab:keyword_validation}
\end{table}

\subsection{Human Validation}

To further validate the selected keywords, they were reviewed by a legal domain expert with professional experience in Sri Lankan legal practice. The objective of this evaluation was to assess whether the keywords represent meaningful legal concepts commonly used in Sinhala legal discourse.

The expert confirmed the relevance of the majority of the selected keywords and their association with legal concepts. The review also identified two keywords, \hspace*{-3pt}\raisebox{-0.5ex}{ \includegraphics[height=1.3\fontcharht\font`\A]{Sinhala_Words/si_word_balaya.pdf}} and \hspace*{-3pt}\raisebox{-0.5ex}{ \includegraphics[height=1.3\fontcharht\font`\A]{Sinhala_Words/si_word_kondesi.pdf}}, as being more strongly associated with constitutional contexts than with the broader body of legal practice. While alternative terms were suggested, these keywords were retained because the corpus frequency analysis demonstrated their substantial presence across the legal document collection.

Overall, this expert review and validation support the legal relevance of the selected keywords, while the corpus frequency analysis provides quantitative evidence of their usage throughout the dataset. Together, these validation methods indicate that the selected keywords represents Sinhala legal text and suitable for legal-domain adaptation.


\end{document}